\documentclass[10pt,twocolumn,letterpaper,numbers]{article}

 \usepackage{cvpr}              % To produce the CAMERA-READY version
\usepackage[ruled,vlined]{algorithm2e}
\usepackage{float}
\usepackage{multicol}
\usepackage[most]{tcolorbox}
    \newtcolorbox{myyellowbox}{
      colback=yellow!20!white, % Light yellow background
      colframe=black, % Black border
      boxrule=0.5pt, % Thin border thickness
      arc=10pt, % Rounded corners (radius)
      outer arc=10pt, % Ensures outer corners are also rounded
      width=\linewidth, % Stays within column width
      parbox=false, % Allows for line breaks within the box
      left=4pt, right=4pt, top=4pt, bottom=4pt % Padding around text
    }

\usepackage[most]{tcolorbox}
    \newtcolorbox{mybluebox}{
      colback=blue!20!white, % Light yellow background
      colframe=black, % Black border
      boxrule=0.5pt, % Thin border thickness
      arc=10pt, % Rounded corners (radius)
      outer arc=10pt, % Ensures outer corners are also rounded
      width=\linewidth, % Stays within column width
      parbox=false, % Allows for line breaks within the box
      left=4pt, right=4pt, top=4pt, bottom=4pt % Padding around text
    }

\usepackage{siunitx}
\definecolor{cvprblue}{rgb}{0.21,0.49,0.74}
\usepackage[pagebackref,breaklinks,colorlinks,allcolors=cvprblue]{hyperref}

\def\paperID{xxxxx} % *** Enter the Paper ID here
\def\confName{CVPR}
\def\confYear{2026}
\usepackage{fontawesome6}

\usepackage{gradient-text}

\title{The Unwritten Benchmark: A New Challenge for Multimodal Machine Learning in Abstract Perceptual Reasoning}

\author{
\textbf{Garima Arya Yadav},
\textbf{Nilay Yilmaz},
\textbf{Yezhou Yang}\\[0.4em]
Arizona State University\\[0.3em]
\texttt{\small \{arya.yadav, nyilmaz3, yz.yang\}@asu.edu}\\[0.25em]
{\small
\centering
\href{https://riri-y.github.io/unwritten-benchmark/}{\color{blue}\tt\small The Unwritten Benchmark Project Page}
}
}
\begin{document}
\maketitle
\begin{abstract}
Current multimodal models have demonstrated remarkable proficiency in recognizing static visual and auditory content. However, their capacity for abstract perceptual reasoning, inferring unseen information from dynamic, generative processes, remains a critical and underexplored frontier. In this paper, we introduce The Unwritten Benchmark, a new challenge designed to probe this abstract perceptual and cognitive ability. We define the core task as acousto-kinematic word inference: models must decipher words, across 3 different writing styles, being written solely from the audio of pen scratches and the video of hand movements, without any visible ink trace. Our evaluation results reveal a profound gap between human and machine performance: while human participants achieve high ordered letter accuracy (over 80\%), leading Multimodal Machine Learning Models, including GPT-4o and Gemini 2.5-Pro, struggle significantly, failing to surpass 10\%. Furthermore, we identify a paradoxical fusion effect in the models, where providing both modalities often degrades performance rather than improving it. This finding indicates a fundamental breakdown in their ability to synthesize complementary perceptual cues for this cognitive task. These findings highlight significant limitations in both cross-modal causal reasoning and the understanding of the micro-kinematics essential for such cognitive and intuitive perceptual reasoning.
\end{abstract}    
\section{Introduction}

The field of multimodal machine learning has made remarkable strides, largely by developing models that are capable of accomplishing perception. These models have achieved strong performance on tasks like Visual Question Answering (VQA)\citep{Antol_2015_ICCV} and image captioning \citep{vinyals2015show}, where the primary challenge is to correlate explicit, static features in one modality (an image of a dog) with corresponding concepts in another (the word "dog"). This success is built on a foundation of recognizing, describing, and linking content that co-occurs across different sensory inputs. While powerful, this paradigm primarily tests a model's performance on recognition and description tasks, correlating descriptive features that are already explicitly present in the data.

However, a deeper and more quintessentially human form of perception involves the "perception of becoming": the ability to observe a dynamic, generative process and infer the abstract outcome it is creating \cite{Johansson1973VisualPO}. This cognitive ability moves beyond static correlation, as deducing an abstract result from a dynamic physical process requires reasoning about causality, movement, and intent. It directs a model to synthesize a constellation of subtle, temporal cues from different senses to understand what is being created, even when the final product is itself unseen. Humans perform this intuitive reasoning constantly; we infer a speaker's intent from subtle vocal shifts and body language, or understand the goal of an action by combining the sight of preparatory motions with their associated sounds \cite{Ambady1992ThinSO}.

We argue that this critical capability remains a significant blind spot in modern AI. This gap persists primarily due to the absence of benchmarks designed to isolate and measure this specific perceptual-cognitive skill. Existing benchmarks excel at testing the recognition of explicit content but fail to probe a model's ability to infer an abstract concept from the physical process of its creation. Consequently, this profound gap between current machine perception and human process-oriented intuition remains largely unaddressed.

\begin{figure*}[t!]
    \centering
    \includegraphics[width=0.93\linewidth]{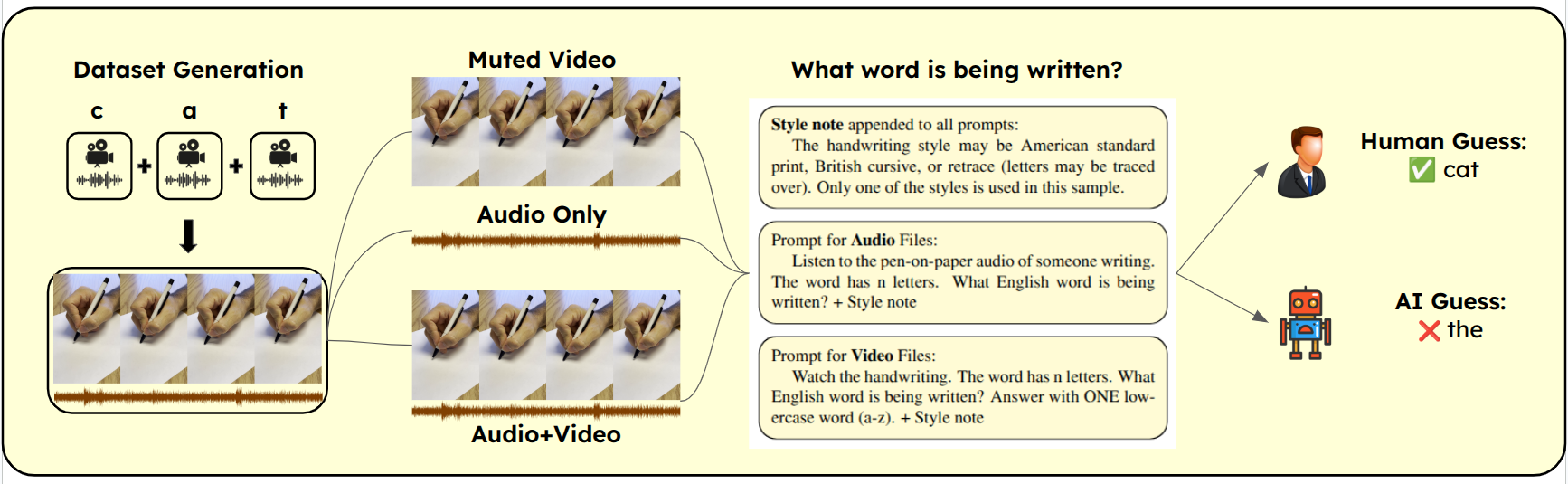}
    \caption{Overview of The Unwritten Benchmark. Words are synthesized by combining audio-visual recordings of individual letters. Each sample is presented in three modalities: Muted Video, Audio Only, and Audio+Video (fused audio). Prompts are identical across video and audio+video; only the input differs. Humans reliably identify the written word, while MLLMs often fail.}
    \label{fig:placeholder}
\end{figure*}

To address this limitation, we introduce The Unwritten Benchmark, a novel diagnostic task that shifts the challenge from static recognition to dynamic, inferential reasoning. The acousto-kinematic word inference task challenges models to decipher a word being written in the complete absence of a visible ink trace. The only information provided is the real-time kinematics of the hand's motion (video) and the corresponding auditory feedback of the pen scratching on paper (audio). This task, which is cognitively intuitive for humans, proves to be profoundly difficult for current state-of-the-art (SOTA) Multimodal Large Language Models (MLLMs).
Our key contributions are as follows:
\begin{itemize}
\item A Novel Multimodal Benchmark: We present a novel dataset containing synchronized audio and video recordings of three handwriting styles without visible ink, creating a unique challenge for abstract multimodal perception.

\item A Scalable Dataset Generation Framework: We detail a methodology for creating a large-scale, robust word dataset by programmatically concatenating a foundational library of high-quality, live-recorded letter primitives. This modular design ensures consistency and allows researchers to generate new word lists or even expand the benchmark to other languages using the Latin alphabet.

\item Revealing a Significant Limitation in Models: We demonstrate that state-of-the-art models, including GPT-4o and Gemini 2.5-Pro, consistently and categorically fail at this task. The best models achieve ordered letter accuracies below 10\%, highlighting a significant gap in their reasoning capabilities compared to human performance, which exceeds 80\%.

\item Identifying a Paradox in Multimodal Fusion: We provide evidence that for this acousto-kinematic task, the fusion of audio and video modalities does not consistently improve AI performance and can even be detrimental. This "paradoxical fusion effect" suggests that current architectures are ill-equipped to synthesize complex, causally-linked temporal cues.

\end{itemize}

\section{Related Work}

\textbf{Handwriting Recognition Benchmarks and Limitations}: Offline handwriting datasets such as IAM database \cite{marti2002iam}, NIST \cite{cohen2017emnist}, and EMNIST \cite{grother1995nist} consist of static images of written text for optical character recognition. These corpora have been instrumental for evaluating handwriting recognition, writer identification, and word spotting \cite{hussain2015comprehensive} tasks, but they treat writing as a static image rather than a dynamic process. In contrast, online handwriting datasets like IAM-OnDB \cite{Liwicki2005IAMOnDBA} and the UNIPEN  project \cite{576870} capture the pen’s trajectory over time, recording sequences of coordinates with timing and pen-up/pen-down events \cite{hussain2015comprehensive}. Such temporal cues can make recognition easier by revealing stroke order and speed \cite{Hanif2023StrokesTR}. However, trajectory-based models assume full access to the pen path and rely on specialized hardware such as digital tablets or motion sensors, which is impractical in many real-world settings. Their performance degrades sharply if stroke data is noisy, partial, or unavailable. Moreover, purely trajectory-driven approaches ignore other cues that humans use – for example, the scratching sound of the pen on paper or contextual scene clues. Some nascent work has started exploring these alternative signals: for instance, the WritingHacker system \cite{yu2016writinghacker} showed that pen-on-paper audio captured by a smartphone can be used to partially infer written text . While promising, such side-channel approaches remain preliminary. In summary, existing handwriting benchmarks do not address scenarios where the writing must be understood without any visible trace of ink. Our benchmark provides the first challenge of this kind, requiring models to infer the written word solely from the indirect, acousto-kinematic cues of the hand's motion and the pen's sound, completely removing the explicit visual trace.

\textbf{Multimodal Action and Sound Recognition:}
Beyond handwriting, many multimodal datasets combine visual and auditory streams to recognize actions. Video benchmarks like Kinetics \cite{carreira2017quo} and EPIC-Kitchens \cite{damen2018scaling} offer clips of everyday activities (often with corresponding audio) for action classification. Large curated audio collections such as AudioSet \cite{gemmeke2017audioset} cover hundreds of sound event classes, and audio-visual sets like VGG-Sound \cite{chen2020vggsound} pair video clips with their synchronized sound. More specialized datasets provide richer context: Ego4D \cite{grauman2022ego4d} is an egocentric video corpus with audio supporting tasks like episodic memory recall and prediction, CREMA-D \cite{cao2014crema} contains face videos and voice clips annotated with emotions, and recent datasets like VGG-Sound Source \cite{chen2021localizing} and STARSS23 \cite{shimada2023starss23} add spatial audio-visual annotations to identify where sounds originate. These benchmarks have advanced multisensory learning, but their tasks involve mapping observable cues directly to labels. In contrast, our work introduces a novel task by shifting the challenge from recognizing observable events to inferring an abstract, symbolic outcome, the word, which is never itself visible or audible.

\textbf{Multimodal Fusion Strategies and Challenges:}
To exploit multiple modalities, researchers have developed various fusion strategies. Early fusion integrates modalities at the feature level, concatenating audio and visual features before feeding to a model, whereas late fusion combines modality-specific predictions at the decision level \cite{baltrusaitis2018multimodal}. 
More sophisticated are cross-modal or mid-fusion approaches \cite{Lu2019ViLBERTPT}, where modalities interact through
attention mechanisms or co-training at intermediate layers. However, a common issue across fusion methods is their reliance on statistical correlation
which often makes the models latch onto coincident patterns present in training data
instead of learning why signals co-occur \cite{Geirhosetal20}. For example, a multimodal model might erroneously predict an action
just because a characteristic sound is present, even if the sound’s source is unrelated in a given scene. Such
failures indicate the model learned “X and Y often occur together” but not “X is caused by Y.” These static
correlations can bias models toward superficial cues \cite{liyicong2025causality}. Indeed, standard fusion strategies, whether early feature integration or late prediction merging, often encourage learning of coincident patterns rather than genuine cross-modal reasoning \cite{baltrusaitis2018multimodal,liyicong2025causality}. As a result, models may latch onto superficial audiovisual cues and falter when deeper inference is required. Notably, no existing audio-visual benchmark forces models to infer an unobserved event or state from the given modalities. The Unwritten Benchmark is designed to fill this crucial gap, challenging models to move beyond statistical correlation and perform causal synthesis by inferring a hidden, abstract product (the word) from its two different, subtle, and causally-linked manifestations (motion and sound).

\section{Dataset and Methodology}

Our work, The Unwritten Benchmark, is centered on a novel, meticulously constructed multimodal dataset designed to rigorously test abstract reasoning in AI (see Figure \ref{fig:placeholder}). It consists of synchronized audio and video recordings of handwriting with no visible ink trace. The dataset creation process, detailed below, ensures consistency, integrity, and reproducibility.

\subsection{Data Collection} \label{sec3.1}

The foundation of our dataset is a collection of live-recorded, single-letter capital and lowercase samples. We recruited three participants, each with a distinct and consistent handwriting style, following standard stroke patterns for the American standard alphabet and British cursive. These stroke patterns, followed by the participants for writing the alphabet, were copied from standard alphabet tracing worksheets provided to American and British school students during their elementary years (see Appendix A.4). %(see Appendix A.4.). 
For each of the first two participants, we recorded all 26 uppercase and lowercase letters of the English alphabet in their preferred writing style from either American Standard or British Cursive. The third participant contributed a specialized 'Retrace' set, consisting of 17 letters that have high stroke variance and repetition of strokes, such as 'a', 'o', 'i', 'p,' to specifically test model performance on common handwriting outliers and special writing styles as well. In summary, across three modalities, Audio, Muted Video, and Synced Audio+Video, the following writing styles were collected:
\begin{enumerate}
    \item \textbf{Standard:} The conventional, non-connected letter style.
    \item \textbf{Cursive:} The classic, flowing, connected style.
    \item \textbf{Retrace:} A style where strokes are intentionally re-traced, representing common handwriting habits and outliers.
\end{enumerate}

Crucially, the recordings were made with a dry pen on a blank white sheet of paper, ensuring that the visual modality contained only the kinematics of the hand's motion, with no visible ink. The audio and video were captured simultaneously using a high-resolution camera and a high-fidelity microphone placed close to the pen nib. To achieve sub-millisecond precision, each audio and video file pair from the live recordings was manually synced, a process that guarantees the integrity of our base multimodal data. This resulted in a foundational set of 409 individual letter-level files, forming the building blocks for our synthetic word dataset.

In addition to the letter primitives, we also recorded a smaller, fully natural set of complete words to serve as a validation set for our synthetic data and for future analysis of natural co-articulation effects. This resulted in a collection of 309 live-recorded word files, also spanning all three styles and modalities.

\subsection{The Synthetic Unwritten Benchmark Dataset}

To create a robust and scalable word-level benchmark, we developed a semi-synthetic dataset by programmatically concatenating our live-recorded letter clips. This approach allowed us to generate a large-scale, consistent word-level dataset from a smaller set of high-quality, live-recorded primitives.

Let $\mathcal{L} = \{'a', 'b', \dots, 'z'\}$ be the set of all lowercase English letters.

Let $\mathcal{S} = \{ \text{Standard, Cursive, Retrace} \}$ be the set of writing styles.

Let $\mathcal{M} = $ \{Audio (A), Muted Video (MV), Audio+Video (AV)\} be the set of modalities.

For each style $s \in \mathcal{S}$ and each letter $l \in \mathcal{L}$, we have a corresponding set of live-recorded media files $F_{s,l} = \{ f_{s,l}^A, f_{s,l}^{MV}, f_{s,l}^{AV} \}$.

Thus, the set of available letters for a given style $s$ is:
$$
\mathcal{L}_{s}^{\text{avail}} = \{l \in \mathcal{L} \mid \forall m \in \mathcal{M}, \text{file } f_{s,l}^m \text{ exists}\}.
$$

\subsubsection{Word Selection and Filtering Algorithm}

Our word list for each style was generated by filtering a comprehensive list of common English words from wordfreq library \cite{robyn_speer_2022_7199437}, $W_{\text{raw}}$, based on the availability of corresponding letter clips (see Algorithm \ref{alg:word_list_gen_2}). Each word was constrained to contain only lowercase English letters (a-z) and have a length ranging from 2 to 5 letters. For each of the three writing styles, we only included words for which all constituent letters had a corresponding live-recorded clip in all three modalities. This was particularly important for the 'Retrace' style, which does not contain all 26 letters due to its specialized nature. Words were composed exclusively of letters from the same writing style, such that a 'Standard' word uses only 'Standard' letter clips, ensuring a consistent style throughout each sample. This process yielded a curated list of words for which a valid, tri-modal sample could be synthesized.

\subsubsection{Concatenation and Synthesis Algorithm}

We created each word's three modalities (audio, muted video, and video with audio) by concatenating the corresponding per-letter clips in their correct sequence. This was done using a robust pipeline to preserve the original media properties and prevent corruption. The final word samples are created by sequentially concatenating the media files of the individual letters. For a word $w$ composed of the sequence of letters $l_1l_2...l_n$, the process is defined in Algorithm \ref{alg:multimodal_synth_2}.

This methodology guarantees that the final word samples are composed entirely of consistent, style-specific letter clips, preserving their original kinematic and auditory properties without re-encoding. This automated process is highly expandable, allowing for the generation of a vast number of word samples for various use cases, including model training. For the benchmark evaluation in this paper, we generated a specific test set totaling \textbf{10,491} files.

\begin{algorithm}
\SetAlgoLined
\DontPrintSemicolon
\KwData{$W_{\text{raw}}$: a list of common English words; $\mathcal{L}_{s}^{\text{avail}}$: the set of available letters for each style $s \in \mathcal{S}$.}
\KwResult{$W_s$: a list of valid words for each style $s$.}
\Begin{
  Initialize $W_s = \emptyset$ for all $s \in \mathcal{S}$.\;
  \For{each word $w \in W_{\text{raw}}$}{
    Convert $w$ to lowercase.\;
    \If{($2 \leq |w| \leq 5$) and ($w \in [a-z]^*$)}{
      \For{each style $s \in \mathcal{S}$}{
        \If{($\forall c \in w, c \in \mathcal{L}_{s}^{\text{avail}}$)}{
          Add $w$ to the list $W_s$.\;
        }
      }
    }
  }
  \Return $W_s$ for each style $s$.\;
}
\caption{Word List Generation}
\label{alg:word_list_gen_2}
\end{algorithm}

\begin{algorithm}
\SetAlgoLined
\DontPrintSemicolon
\KwData{$w = l_1l_2...l_n \in W_s$: a word from a style-specific word list; $F_{s,l}^m$: a set of available letter clips.}
\KwResult{A multimodal word sample $W_{w,s} = \{ w_{w,s}^A, w_{w,s}^{MV}, w_{w,s}^{AV} \}$.}
\For{each modality $m \in \mathcal{M}$}{
  Initialize an empty output stream $w_{w,s}^m$.\;
  \For{$i = 1$ to $n$}{
    Get the source media file for the $i$-th letter: $f_{\text{source}} = f_{s, l_i}^m$.\;
    Concatenate $f_{\text{source}}$ to the end of the output stream $w_{w,s}^m$.
  }
}
\Return $W_{w,s}$.\;
\caption{Multimodal Word Synthesis}
\label{alg:multimodal_synth_2}
\end{algorithm}

\subsection{Dataset Extensibility and Language Agnosticism}

The design of The Unwritten Benchmark makes it highly extensible and language-agnostic. The dataset's foundation is a library of individual, live-recorded letter primitives. Because the word-level samples are synthetically generated from this library, the benchmark is not limited to a fixed vocabulary.

The described concatenation methodology allows for the programmatic creation of new word lists and sentences. This enables the generation of millions of unique data samples in English. Furthermore, the existing foundational clips for the Latin alphabet can be used to generate vast amounts of data for any language sharing this alphabet, such as French, Spanish, or German, without any additional effort towards data collection.

By utilizing the automated dataset generation pipeline, the benchmark can be expanded to entirely new writing systems, such as Cyrillic or Arabic, simply by adding new live-recorded character primitives. This is because the core task of inferring symbols from their physical, acousto-kinematic creation process, is a concept that applies equally to any writing system. This modular and scalable design ensures the benchmark can evolve to foster cross-linguistic research and provide a sustainable resource for years to come.

\section{Experimental Setup}\label{setup}

Our experimental design aims to provide a rigorous and reproducible evaluation of state-of-the-art multimodal models on The Unwritten Benchmark.

\subsection{Models and Baseline}

We evaluated four leading MLLMs in a zero-shot setting to test their intrinsic, non-task-specific reasoning: GPT-4o\citep{openai2024gpt4o}, Gemini 2.5 Pro\citep{google2024gemini}, Gemini 2.5 Flash\citep{google2024gemini}, and Qwen2.5-Omni-7B\citep{qwen2024qwen2.5}.

Specific implementation details were required for two models. For GPT-4o, Audio+Video (AV) evaluation could not be performed, as its current API does not accept pre-merged AV files; these results are therefore omitted. For Qwen2.5-Omni-7B, we implemented a graceful backtracking mechanism with progressively lower frame rates, which gradually decreases from 30 fps down to 10 fps, to ensure robust processing of longer videos that occasionally failed at the standard 30 fps.

To contextualize the performance of the models, we also established a Human Baseline, which serves as a practical upper bound for the task.

\subsection{Prompting Strategy}

Initial exploratory experiments, as shown in Appendix A.3, revealed that open-ended prompts resulted in models frequently defaulting to common, short words, such as "the" and "cat", a form of mode collapse. To elicit more meaningful responses and conduct a more rigorous evaluation, we incorporated a length constraint into the prompt. After significant prompt engineering, the variable {n} was programmatically populated with the correct word length for each sample for all evaluations.

\subsection{Evaluation Metric}

The full word accuracy was near-zero for all evaluated models. This indicates a near-total failure to solve the task outright. To provide a more granular and informative signal of partial understanding, we adopted Ordered Letter Accuracy (OLA) as our primary metric. This metric measures the percentage of characters in the model's guess that match the ground truth character in the correct sequence, offering a more nuanced view of performance than a strict binary score.

To calculate OLA, let $W_{true}$ be the ground truth word and $W_{pred}$ be the predicted or guessed word, both of length $n$. The OLA is calculated as:$$\text{OLA}(W_{true}, W_{pred}) = \frac{1}{n} \sum_{i=1}^{n} \mathbb{I}(W_{true}[i] = W_{pred}[i])$$where $\mathbb{I}$ is the indicator function, evaluating to 1 if the characters at position $i$ match, and 0 otherwise. For example, if the correct word is "cat" and the model guesses "bot", only the 't' at the third position matches. The OLA is therefore $1/3 \approx 0.33$.

\subsection{Human Evaluation}

The primary task for both human and AI participants was to infer the English word or letter being written from the provided multimodal cues. The evaluation corpus consisted of a small subset of words, randomly sampled from our generated word lists while ensuring a balanced distribution across styles and word lengths, resulting in 300 total files for evaluation.

To establish a gold standard for performance, we conducted a human evaluation study with 20 adult participants, all of whom were native or fluent English speakers. The study was hosted on Amazon Mechanical Turk (MTurk) and designed to rigorously measure human perception. Participants were presented with trials containing the three aligned modalities (Audio, Muted Video, and Audio+Video) for some word or letter and were tasked with typing the word they perceived. Each participant completed 15 trials, with trials rotating across participants to ensure comprehensive coverage of the evaluation corpus.

    \begin{table}
    \caption{Overall Ordered Letter Accuracy across primary modalities.}
    \label{tab:modality-results}
    \centering
    \small
    \begin{tabular}{l S[table-format=2.2] S[table-format=2.2] S[table-format=2.2]}
    \toprule
    \multicolumn{1}{c}{\textbf{}} &
    \multicolumn{3}{c}{\textbf{Modality}} \\
    \cmidrule(lr){2-4}
    & \multicolumn{1}{c}{\textbf{Audio}} 
    & \multicolumn{1}{c}{\textbf{Muted Video}} 
    & \multicolumn{1}{c}{\textbf{Audio+Video}} \\
    \midrule
    \textbf{Human} & \textbf{{19.47}} & \textbf{80.78} & \textbf{77.01} \\
    \midrule
    Qwen2.5 Omni    & 7.83  & 5.03  & 3.95  \\
    GPT-4o          & 8.87  & 8.85  & {--}  \\
    Gemini 2.5 Pro  & 10.04 & 8.71  & 9.07  \\
    Gemini 2.5 Flash& 9.49  & 8.38  & 8.44  \\
    
    \bottomrule
    \end{tabular}
    \end{table}

\section{Results and Analysis}

\subsection{Observations and The Paradoxical Fusion Effect}

We evaluated the performance of current MLLMs on the Unwritten Benchmark. Table \ref{tab:modality-results} presents their accuracy evaluated by the OLA metric. Current state-of-the-art models consistently demonstrate poor performance in all modalities, with results falling below 10\%. This indicates a categorical failure to solve the task. The best-performing model, Gemini 2.5-Pro, fails to surpass even 11\% accuracy on any single task, with its peak score (10.04\%) coming from the audio-only modality. All other models lag behind, with scores often hovering in the single digits and, in the case of Qwen2.5-Omni, dropping as low as 3.95\%.

%    \begin{figure}[H]
 %       \centering
  %      \includegraphics[width=1\linewidth]{graph_overall_letter_accuracy.png}
   %     \caption{Overall Ordered Letter Accuracy across primary modalities (rounded to one decimal).}
    %    \label{fig:placeholderfig1}
    %\end{figure}

\begin{table*}[t]
\caption{Ordered Letter Accuracy by Handwriting Style and Modality (A: Audio, MV: Muted Video, AV: Audio+Video).}
\label{tab:style-modality-results2}
\centering
\small
\begin{tabular}{
    l
    S[table-format=2.2] S[table-format=2.2] S[table-format=2.2]
    S[table-format=2.2] S[table-format=2.2] S[table-format=2.2]
    S[table-format=2.2] S[table-format=2.2] S[table-format=2.2]
}
\toprule
\multicolumn{1}{c}{\textbf{Method}} &
\multicolumn{3}{c}{\textbf{Standard}} &
\multicolumn{3}{c}{\textbf{Cursive}} &
\multicolumn{3}{c}{\textbf{Retrace}} \\
\cmidrule(lr){2-4} \cmidrule(lr){5-7} \cmidrule(lr){8-10}
& \multicolumn{1}{c}{\textbf{A}} 
& \multicolumn{1}{c}{\textbf{MV}} 
& \multicolumn{1}{c}{\textbf{AV}} 
& \multicolumn{1}{c}{\textbf{A}} 
& \multicolumn{1}{c}{\textbf{MV}} 
& \multicolumn{1}{c}{\textbf{AV}} 
& \multicolumn{1}{c}{\textbf{A}} 
& \multicolumn{1}{c}{\textbf{MV}} 
& \multicolumn{1}{c}{\textbf{AV}} \\
\midrule
\textbf{Human}   & \textbf{24.60} & \textbf{84.56} & \textbf{87.58} 
                 & \textbf{21.02} & \textbf{65.57} & \textbf{57.22} 
                 & \textbf{13.09} & \textbf{93.94} & \textbf{91.55} \\
\midrule
Qwen2.5 Omni     & 7.65 & 5.10 & 3.54 & 7.69 & 4.26 & 4.03 & 8.91 & 7.37 & 5.08 \\
GPT-4o           & 8.92 & 8.66 & {--} & 8.83 & 9.05 & {--} & 8.79 & 8.78 & {--} \\
Gemini 2.5 Pro    & 10.04 & 8.87 & 8.75 & 9.56 & 8.65 & 9.32 & 11.69 & 8.41 & 9.27 \\
Gemini 2.5 Flash  & 9.22 & 8.61 & 8.49 & 9.54 & 8.37 & 8.41 & 10.26 & 7.67 & 8.42 \\
\bottomrule
\end{tabular}
\end{table*}
This categorical failure is thrown into sharp relief when contrasted with the Human Baseline. While models struggle, human participants demonstrate a high capacity for the task, achieving 80.78\% accuracy on Muted Video (MV) and 77.01\% on Audio+Video (AV). The human results prove that the kinematic information in the video is a rich, solvable, and primary source of information. The models' inability to capitalize on this, combined with the human's mastery of it, signifies that the cognitively intuitive, abstract perceptual reasoning humans employ is fundamentally absent in these SOTA architectures.

Furthermore, a closer analysis of the modalities reveals a ``paradoxical fusion effect'' in the AI models. For Qwen2.5-Omni, providing both audio and video data (3.95\%) significantly causes a lower score than processing either only audio (7.83\%) or only video (5.03\%) input. The Gemini models, while not exhibiting a negative effect, show no meaningful benefit from the combined Audio + Video input, with AV scores (9.07\% for 2.5 Pro and 8.44\% for 2.5 Flash) that are nearly identical to their single-modality scores. This failure, where adding more sensory data fails to improve, or even degrades, performance, suggests a fundamental breakdown in the models' fusion architectures. Rather than synthesizing complementary causal signals, the models appear to be confused by the additional information.

This contrasts sharply with the Human Baseline, which demonstrates a more nuanced and logical use of the modalities. While humans also find the audio-only signal challenging (19.47\%), this is by far the most difficult sensory modality for them. This situation is notably different for the AI models, for which audio is often their highest-performing modality. In contrast, human performance soars on Muted Video (80.78\%). Interestingly, the combined Audio+Video (77.01\%) results in a slight overall performance dip. A closer look at Table \ref{tab:style-modality-results2} explains this: for the `Standard' style, audio helps (84.56\% $\rightarrow$ 87.58\%), suggesting humans can use the scratching sound to reinforce the visual strokes. However, for the more ambiguous 'Cursive' style, the audio appears to act as a distraction, causing performance to drop (65.57\% $\rightarrow$ 57.22\%). Our evaluation thus demonstrates a sophisticated, context-dependent fusion strategy is entirely lacking in the AI models and reveals a stark contrast between humans’ intuitive perception and the capabilities of current MLLMs, highlighting significant limitations in their abstract and multimodal reasoning.

\subsection{Qualitative Analysis of Failure Modes: Mode Collapse vs. Signal Interpretation}

Beyond quantitative scores, the nature of the models' errors reveals different failure modes.

\begin{itemize}
    \item \textbf{Mode Collapse:} GPT-4o and Qwen frequently defaulted to outputting common, high-frequency English words such as "the", "pen", "house", and "apple", regardless of the input's kinematic or acoustic complexity. This reveals a complete failure to extract meaningful features from the input, causing the model to retreat to its base textual priors.
    \item \textbf{Attempted Signal Interpretation:} In contrast, the Gemini models produced a much wider and more randomized set of incorrect guesses. Their outputs, while wrong, often seemed to correspond loosely to the input's characteristics. This suggests that these models were at least attempting to interpret the input signal, even if they ultimately failed at the high-level task of mapping it to the correct symbolic concept or letter. 
This reveals a hierarchy of failure, from complete signal rejection to flawed interpretation.
\end{itemize}

\subsection{Analysis Across Handwriting Styles}

A fine-grained analysis of performance across the different handwriting styles in Table \ref{tab:style-modality-results2} reveals divergent strategies and capabilities between human and machine perception. Human performance is exceptionally strong on visual modalities for Standard (84.56\% MV) and Retrace (93.94\% MV) styles, yet drops for the more ambiguous Cursive style (65.57\% MV). 

Qualitative feedback from participants provides insight into these results. Several noted that visual inference was easier for styles they personally used, suggesting a reliance on familiar, internalized motor programs. This was further evidenced by our observation that many participants instinctively mimicked the writing motions with their own fingers; a form of embodied simulation used to better understand the kinematic patterns. One participant highlighted why the Retrace style was visually the easiest, stating, "...the sample would show the person going over strokes twice...which gave us more context and time to understand what was being written." This indicates that humans leverage repetition for contextual reinforcement and temporal clarity.

Interestingly, the MLLMs' performance, while universally low, mirrors the human trend by also peaking on the Retrace style. For example, Gemini 2.5 Pro's audio-only score jumps from 10.04\% (Standard) to 11.69\% (Retrace). This suggests that, unlike humans who use repetition for high-level conceptual clarity, the models are likely latching onto simpler, low-level statistical features. The repetitive motion paths and consistent sound patterns exaggerated in the Retrace style provide a stronger, less noisy signal, but this minor success fails to generalize to the more complex and varied motions of conventional writing.

This divergence in reasoning highlights a fundamental difference in approach. Human participants engaged in a holistic, top-down reasoning process, treating the video as a continuous stream of motion and leveraging intuitive, generative models of kinematics. Their feedback and behavior show they perceive the task as one of high-level, abstract inference. The models, however, appear stuck in a low-level, feature-extraction paradigm. This reinforces the hypothesis that, unlike humans, the models lack a higher-level, conceptual understanding of the generative grammar of motion and the causal link between physical action and intent \cite{LONGCAMP2006681}.

\subsection{Validating the Synthetic Data Generation}

A core component of our methodology is the semi-synthetic generation of the dataset by concatenating live-recorded letter primitives. To ensure this process did not introduce confounding artifacts or negatively impact the validity of our findings, we conducted a validation study. For this study, we evaluated the same state-of-the-art models on the smaller, fully natural set of 309 live-recorded word files described in Section \ref{sec3.1}. These words were recorded in single, continuous takes by our participants and were not synthetically concatenated.
The experimental setup, including the prompting strategy with the word length constraint, remained identical to the main evaluation described in Section \ref{setup}. The goal of this study was to determine if model performance on this live data was commensurate with the performance on our larger, synthetic dataset.

The results of this validation, presented in Table \ref{tab:modality-resultscompar}, strongly validate our data generation approach. The performance of all models on the live-recorded word set is highly consistent with their performance on the main synthetic benchmark shown in Table \ref{tab:modality-results}. While the exact scores show minor differences, with some models performing slightly better or worse on the live data, the overall trends of categorical failure and relative model rankings are identical.

\begin{table}[t]
    \caption{Overall Ordered Letter Accuracy across primary modalities on live-recorded whole words.}
    \label{tab:modality-resultscompar}
    \centering
    \small
    \begin{tabular}{l S[table-format=2.2] S[table-format=2.2] S[table-format=2.2]}
    \toprule
    \multicolumn{1}{c}{\textbf{}} &
    \multicolumn{3}{c}{\textbf{Modality}} \\
    \cmidrule(lr){2-4}
    & \multicolumn{1}{c}{\textbf{Audio}} 
    & \multicolumn{1}{c}{\textbf{Muted Video}} 
    & \multicolumn{1}{c}{\textbf{Audio+Video}} \\
    \midrule
    Qwen2.5 Omni    & 6.50  & 7.79  & 3.76  \\
    GPT-4o          & 11.65  & 8.80  & {--}  \\
    Gemini 2.5 Pro  & 13.15 & 9.56  & 6.37  \\
    Gemini 2.5 Flash& 7.80  & 10.86  & 8.06  \\
    \bottomrule
    \end{tabular}
    \end{table}

GPT-4o's Muted Video performance is nearly identical, scoring 8.85\% on the synthetic set and 8.80\% on the live set. Gemini 2.5 Flash shows similar stability on Audio+Video, scoring 8.44\% (Synthetic) vs. 8.06\% (Live). While there are minor variations, such as Gemini 2.5 Pro's audio score (10.04\% Synthetic vs. 13.15\% Live) or Gemini 2.5 Flash's Muted Video score (8.38\% Synthetic vs. 10.86\% Live), the overall trend followed by different modalities is identical.

\begin{figure}
    \centering
    \includegraphics[width=0.85\linewidth]{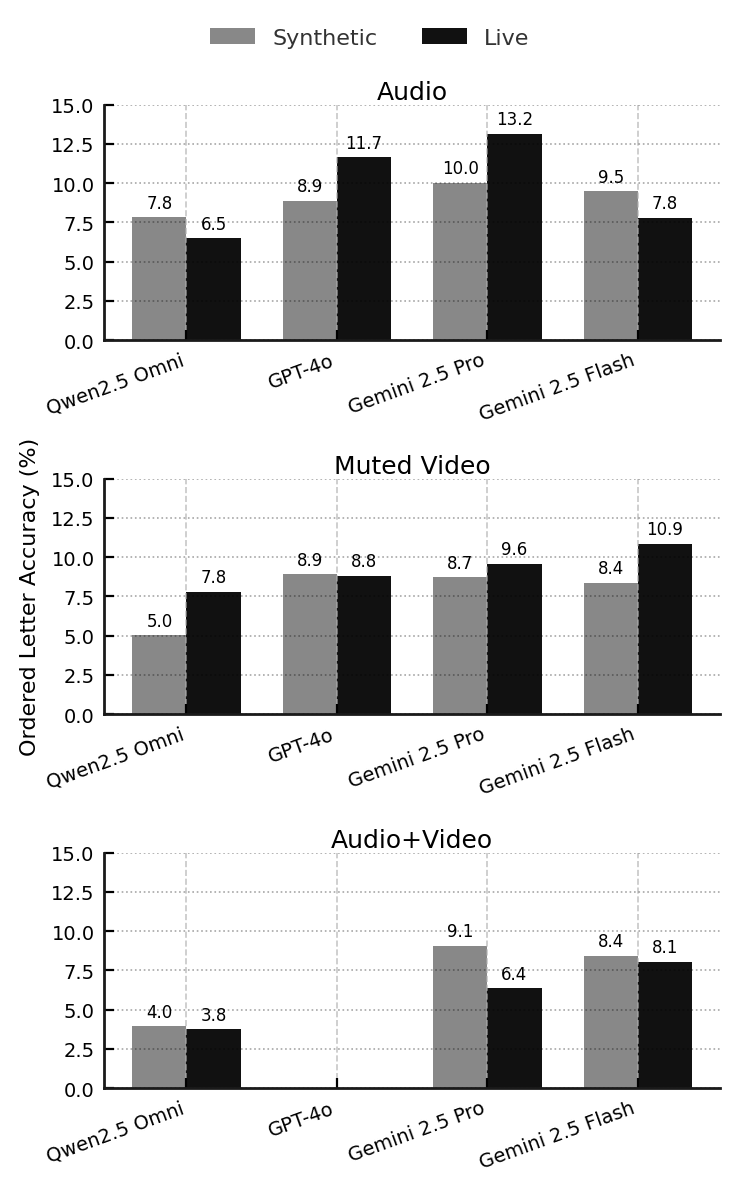}
    \caption{Model accuracy on synthetic vs. live-recorded words shows similar performance.}
    \label{fig:vs}
\end{figure}

As observed in Figure \ref{fig:vs}, across all models and modalities, the accuracy scores remain in a similar range, and the relative performance ranking between models is largely preserved. Furthermore, the paradoxical fusion effect is also still observable. On the live set, Qwen2.5-Omni's performance drops from 6.50\% (Audio) to 3.76\% (Audio+Video), mirroring its drop on the synthetic set (7.83\% to 3.95\%). The congruence of these results demonstrates that our method of concatenating letter primitives to form words does not introduce significant artifacts that would either simplify or complicate the task for the models.

%\begin{figure}
 %   \centering
  %  \includegraphics[width=0.70\linewidth]{output (2).png}
  %  \caption{Model accuracy on synthetic vs. live-recorded words shows similar performance.}
   % \label{fig:vs}
%\end{figure}

\section{Discussion}

The consistent and profound failure of state-of-the-art MLLMs on The Unwritten Benchmark points to a systemic limitation in their reasoning capabilities. Our results suggest that this is not merely a data-domain issue but a deeper architectural and conceptual one. 

\subsection{Beyond Static Frames: The Failure to Perceive Continuous Motion}

A primary reason for the models' failure appears to be their tendency to process videos as a "bag of frames" or a sparse collection of keyframes, rather than as a continuous stream of motion. True perceptual reasoning in this domain requires understanding the flow, velocity, and acceleration of the pen tip, i.e. the micro-kinematics, which is lost when frames are treated in isolation as an image based visual reasoning task. Humans intuitively perceive the continuous path of motion, using it to mentally reconstruct the letterform and infer the word.
Where humans effortlessly recover coherent motional Gestalts from sparse cues \cite{Johansson1973VisualPO}, current deep architectures lack comparable inductive biases and struggle to represent or reason about such continuous motion structure. This is evidenced by the "mode collapse" in some models, which suggests the kinematic and related acoustic information for such abstract task is treated as noise. Future work must focus on developing architectures that can represent and utilize the rich information present in the temporal dynamics of video, moving beyond static-frame analysis toward true spatio-temporal understanding.

\subsection{Toward Causal Fusion: Correlation vs. Causality}

The paradoxical fusion effect highlights the brittleness of current multimodal fusion mechanisms. These systems are typically trained to learn correlations between descriptive, co-occurring features (e.g., the sound of a bark with the image of a dog). However, in our task, the audio of a pen scratch and the video of its movement are not merely correlated; they are two different manifestations of the same underlying causal event, i.e. a pen stroke.

The models' inability to understand this shared cause leads to a breakdown. Instead of using one modality to disambiguate the other, the model appears to be confused by the combined signals, treating them as conflicting evidence rather than complementary parts of a whole. This calls for the development of better fusion techniques that are explicitly designed to model the underlying causal relationships between data streams, a critical step toward more robust and generalizable multimodal reasoning.

\subsection{The Path Forward: Tractability Through Specialization}

The acousto-kinematic word inference task is not computationally impossible. A specialized, heuristic-based system could achieve reasonable performance. For instance, a pipeline using a computer vision library like MediaPipe \citep{lugaresi2019mediapipe} to extract pen, or finger landmark coordinates, followed by a model with a Connectionist Temporal Classification (CTC) \citep{graves2006connectionist}, could plausibly learn to map motion trajectories to character sequences with relative ease. The fact that there exist heuristic ways to solve this problem further isolates the failure of generalist MLLMs: they lack the specific, intuitive reasoning about physics and motion that would allow them to perform this task in a zero-shot setting. This highlights a critical direction for research: imbuing these powerful, general-purpose models with the foundational reasoning capabilities necessary to understand the physical world, rather than relying solely on pattern recognition from massive, static datasets.

\section{Conclusion}

In this paper, we introduced The Unwritten Benchmark, a novel dataset and task designed to evaluate abstract perceptual reasoning in multimodal AI. Our comprehensive evaluation demonstrates that state-of-the-art MLLMs categorically fail at this task, revealing a profound gap between their capabilities and human intuition. We identified a paradoxical fusion effect that challenges common assumptions in multimodal learning, suggesting that current models struggle to reason about dynamic, causally-linked signals.

By defining and validating this clear and currently unsolved problem, we provide a concrete tool for the community to measure progress in a critical area of AI. We hope The Unwritten Benchmark will serve as a catalyst, encouraging a shift in focus from models that merely recognize the world as it is, to models that can understand the physical and causal processes that shape it. We believe this research direction is essential for building more robust, general, and truly intelligent systems.

\section*{Acknowledgments}

NY is supported by the Republic of Türkiye Ministry of National Education.  We thank the NSF NAIRR initiative, the Research Computing (RC) at Arizona State University (ASU) for their generous support in providing computing resources.
The views and opinions of the authors expressed herein do not necessarily state or reflect those of the funding agencies and employers.

%this is empty
%this is empty
{
    \small
    \bibliographystyle{ieeenat_fullname}
    \bibliography{main}
}

\clearpage
\setcounter{page}{1}
\maketitlesupplementary

\appendix
\section{Appendix}
\subsection{Dataset Creation: Detailed Methodology}
The creation of the synthetic \texttt{WORD\_DATASET} involved a two-stage process: (1) generating a clean list of common English words, and (2) programmatically synthesizing the multimodal samples from our live-recorded letter primitives.

\subsubsection{Common Word List Generation}
A high-quality, clean word list was essential for the benchmark's integrity. We started with a high-frequency list of common English words and applied a strict filtering protocol.

\paragraph{Source} The initial seed list was derived from the \texttt{wordfreq} library's top English word lists, which are based on large text corpora. Any similar open-source frequency list can be used to reproduce this process; the critical component is the filtering pipeline.

\paragraph{Filtering Rules} The raw list was cleaned using the following sequential rules:
\begin{enumerate}
    \item \textbf{Length Filter:} Words were restricted to lengths between 2 and 5 characters, inclusive.
    \item \textbf{Character Filter:} Words were required to contain only ASCII letters \texttt{[a-z]} after being converted to lowercase.
    \item \textbf{Stoplist Removal:} We removed words containing digits, hyphens, apostrophes, or other non-alphabetic characters. 
    \item \textbf{Coverage Filter (Per-Style):} A word was excluded from a specific style's word list if any of its constituent letters were missing from that style's set of recorded primitives. This was crucial for ensuring that we could synthesize every word in a given list for that style (e.g., the `Retrace` style has an incomplete alphabet list).
    \item \textbf{Deduplication:} The final lists were deduplicated and stored in their canonical lowercase form.
\end{enumerate}
This process yielded cleaned word lists for each handwriting style, ready for synthesis.

\subsubsection{Word Synthesis Pipeline}
The pipeline maps a word to its constituent letter clips and concatenates them to create the final multimodal sample.

\paragraph{File Naming and Mapping} For each letter `l` in a given style `s`, the system maps to three source files. The file structure is as follows:
\begin{itemize}
    \item \textbf{Standard (s=1):} \texttt{.../l-1.mp3}, \texttt{.../l-1.mp4}, \texttt{.../l-1\_muted.mp4}
    \item \textbf{Cursive (s=2):} \texttt{.../l-2c.mp3}, \texttt{.../l-2c.mp4}, \texttt{.../l-2c\_muted.mp4}
    \item \textbf{Retrace (s=3):} \texttt{.../l-3r.mp3}, \texttt{.../l-3r.mp4}, \texttt{.../l-3r\_muted.mp4}
\end{itemize}
An availability check ensures all three modalities exist for every letter in a word before synthesis proceeds for that word. Styles are never mixed during synthesis. To reduce writer and style bias, the dataset was collected from three distinct participants, each contributing a consistent handwriting style, with each participant recording the style most natural to them.

\paragraph{Concatenation Mechanics} We used a robust \texttt{ffmpeg} pipeline to concatenate clips while preserving media properties.
\begin{itemize}
    \item \textbf{Video Concatenation:} For both muted and audio-visual videos, we used the \texttt{ffmpeg concat demuxer}. This method performs a container-level stitch without re-encoding, making it fast and lossless. A text file listing the absolute paths of the letter clips was generated for each word, and \texttt{ffmpeg} was run with the \texttt{-c copy} flag.
    Example command for creating a video for the word "arm":
    \begin{mybluebox}
        First, create arm.txt:
        \begin{itemize}
            \item file '/path/to/a-1.mp4'
            \item file '/path/to/r-1.mp4'
            \item file '/path/to/m-1.mp4'
        \end{itemize}
        ffmpeg -v quiet -y -f concat -safe 0 -i arm.txt -c copy arm\_final.mp4
        
    \end{mybluebox}
    
    \item \textbf{Audio Concatenation:} To maximize speed and avoid potential decoding errors in source clips, we used a raw bytewise concatenation as the primary method.
    \begin{mybluebox}
        Example command for creating audio for the word "arm"
        
        cat  a-1.mp3  r-1.mp3  m-1.mp3 $>$  arm\_final.mp3
    \end{mybluebox}
\end{itemize}

\paragraph{Timing} No artificial spacing was introduced between letters; the clips are concatenated back-to-back. Future work could explore the impact of inserting fixed-duration silences (audio) and freeze-frames (video) to simulate natural pauses.

\paragraph{Quality Control} To ensure the integrity of the synthesized dataset, every letter primitive was manually verified for synchronization during dataset construction. In addition, random synthesized word samples were manually spot-checked after concatenation to confirm that the stitching process did not introduce noticeable artifacts, corruption, or modality misalignment. These checks were intended to ensure that model behavior reflected the difficulty of the task itself rather than errors in preprocessing or file construction.

\subsection{Human Evaluation: Additional Details}
The human evaluation was designed to establish a robust performance ceiling and gather qualitative insights into how humans solve this task.

\paragraph{Participant Testimonies and Observations}
Beyond the quantitative scores, we collected qualitative feedback and made observations that informed our analysis:
\begin{itemize}
    \item \textbf{Embodied Simulation:} We frequently observed participants instinctively mimicking the writing motions with their own fingers while watching the video clips. This suggests that human perception on this task is not merely passive observation but an active, \textit{embodied simulation} used to better understand and internalize the kinematic patterns. This behavior also appeared in audio-only trials, where participants reported relying on the rhythm and structure of the pen sounds to mentally reconstruct plausible writing motions.

    \item \textbf{Audio-Only Inference Cues: } Human performance in the audio-only condition, while much lower than video-based performance, was still above trivial guessing. This appears to be supported by genuine acoustic structure in the signal, including cues such as stroke count, pauses, timing, and intensity changes that correspond to properties of specific letters or writing patterns. Participant feedback suggested that these rhythmic and temporal cues sometimes enabled partial inference even in the absence of visible motion. 
    
    \item \textbf{Familiarity Bias:} Several participants noted that it was significantly easier to guess words written in a style that matched their own (Standard vs. Cursive). This points to a reliance on familiar, internalized motor programs for inference, and likely helps explain the lower human performance on the British Cursive condition relative to Standard and Retrace.
    
    \item \textbf{Contextual Reinforcement:} One participant provided a particularly insightful comment on the 'Retrace' style: \textit{"it was easiest to guess the retraced words because the sample would show the person going over strokes...which gave us more context and time to understand what was being written."} This confirms our hypothesis that humans leverage repetition for high-level contextual clarity.
\end{itemize}

\subsection{Prompting Strategy}
Initial exploratory experiments revealed that open-ended prompts resulted in models frequently defaulting to common, short words (e.g., "the," "cat"), a form of mode collapse. To elicit more meaningful responses and conduct a more rigorous evaluation, we incorporated a length constraint into the prompt. After significant prompt engineering, the following formats were finalized for all evaluations. The variable {n} was programmatically populated with the correct word length for each sample.

The numeric constraint was used only to restrict the expected output length, not to provide semantic information about the answer itself. In other words, the prompt does not narrow the content of the response beyond the number of letters; it simply prevents degenerate outputs and makes evaluation more comparable across samples.

\textbf{Final Prompts}

\begin{myyellowbox}
    \textbf{Style note} appended to all prompts:
    
    The handwriting style may be American standard print, British cursive, or retrace (letters may be traced over). Only one of the styles is used in this sample.
\end{myyellowbox}

\begin{myyellowbox}
    Prompt for \textbf{Audio} Files: 
    
    Listen to the pen-on-paper audio of someone writing. The word has {n} letters. What English word is being written? Answer with ONE lowercase word (a-z). If unsure, guess the word. Do not apologize. Do not explain. Return only the word." + Style note
\end{myyellowbox}

\begin{myyellowbox}
    Prompt for \textbf{Video} Files: 
    
   Watch the handwriting. The word has {n} letters. What English word is being written? Answer with ONE lowercase word (a-z). If unsure, guess the word. Do not apologize. Do not explain. Return only the word." + Style note
\end{myyellowbox}

The above prompts provided to the MLLMs were finalized after a thorough iterative process designed to minimize ambiguity and prevent common failure modes like mode collapse. Some of the prompts tried and later discarded are as follows:

\paragraph{V1: Initial Open-EndedPrompt - Discarded}
\begin{myyellowbox}
 "What word is being written in this video?"
\end{myyellowbox}
\textbf{Result:} This prompt consistently led to total failure as all models started apologizing and no meaningful results were produced in preliminary tests.

\paragraph{V2: Initial Open-Ended Prompt with Answer length limit - Discarded}

\vspace{5pt}
\begin{myyellowbox}
 "What word is being written in this video?  Answer with ONE lowercase word (a-z). Do not apologize."
\end{myyellowbox}
\textbf{Result:} This prompt consistently led to models defaulting to short, high-frequency words ("the," "a," "is"), regardless of the input's duration or complexity.

\paragraph{V3: Added Style Note - Discarded}

\vspace{3pt}

\begin{myyellowbox}
    "Analyze the handwriting in the video. The style could be standard, cursive, or retrace. What word is being written?"
\end{myyellowbox}

\textbf{Result:} This prompt still suffered from significant mode collapse. The style note alone was insufficient to guide the models toward a more detailed analysis.

\paragraph{V4: Prompt with Length Constraint - Discarded}

\begin{myyellowbox}
    Watch the handwriting. The word has {n} letters. What English word is being written? Answer with ONE lowercase word (a-z).
\end{myyellowbox}

\textbf{Result:} The inclusion of the word length \texttt{\{n\}} was the most critical factor in eliciting meaningful responses, forcing the models to move beyond their default priors and attempt a genuine interpretation of the input signal.

Finally, a combination of V3 and V4 was tried, which yielded the best results so far, and was thus used to form the Final prompts

\subsection{Handwriting Style Guidance}
To ensure consistency in the foundational letter primitives, participants were guided by standard handwriting worksheets commonly used in elementary education. For the 'American Standard' and 'British Cursive' styles, participants referred to worksheets that provided stroke-order diagrams and dotted outlines for each uppercase and lowercase letter.

\begin{figure}
    \centering
    \includegraphics[width=0.75\linewidth]{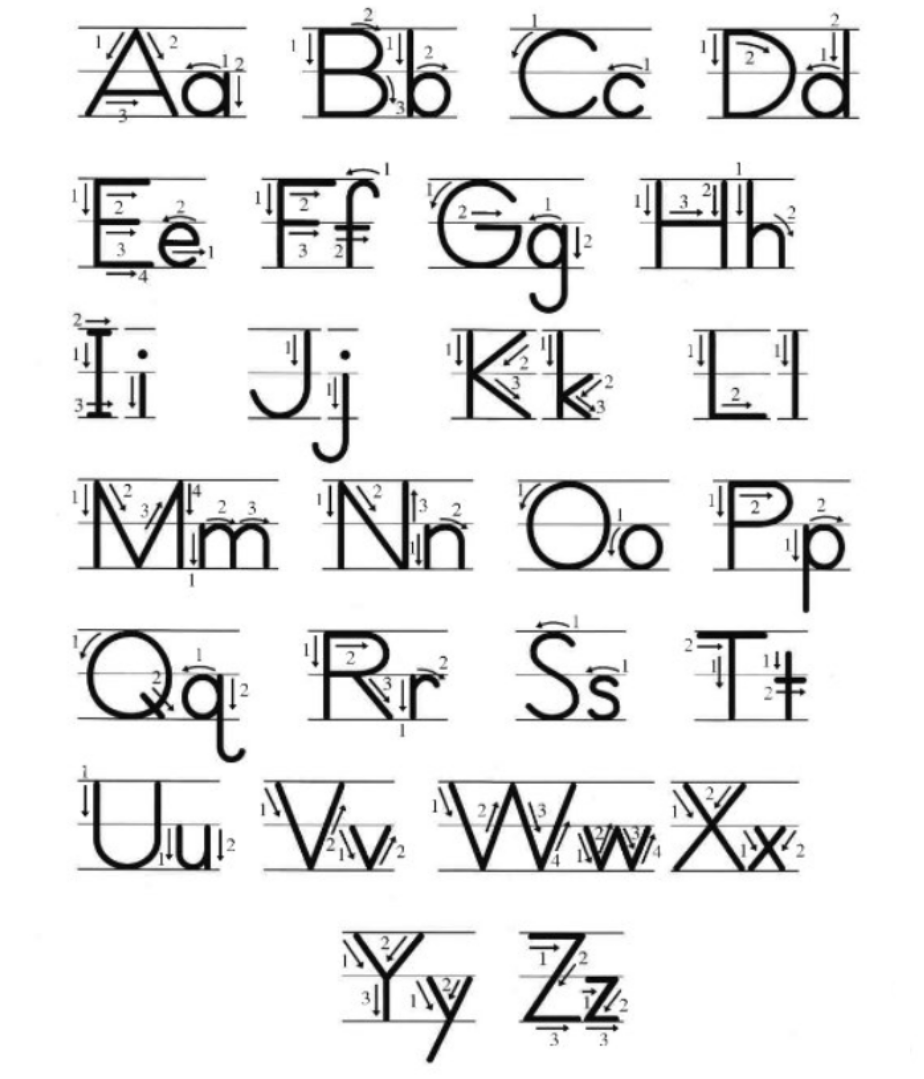}
    \caption{Stroke pattern for American Standard English that was followed during data collection.}
    \label{fig:placeholder}
\end{figure}

\begin{figure}
    \centering
    \includegraphics[width=0.75\linewidth]{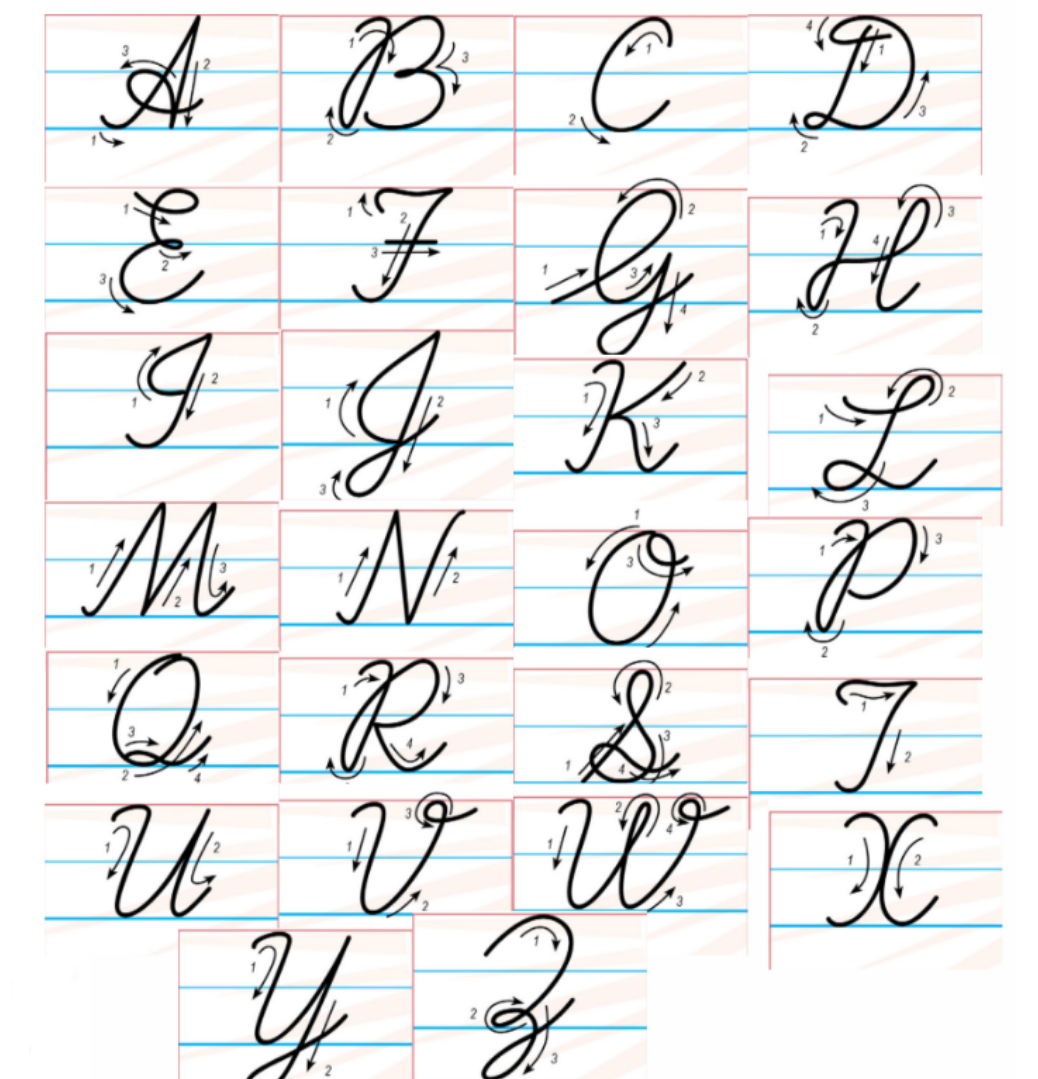}
    \caption{Stroke pattern for British Capital cursive that was followed during data collection.}
    \label{fig:placeholder}
\end{figure}

\begin{figure}[H]
    \centering    \includegraphics[width=1\linewidth]{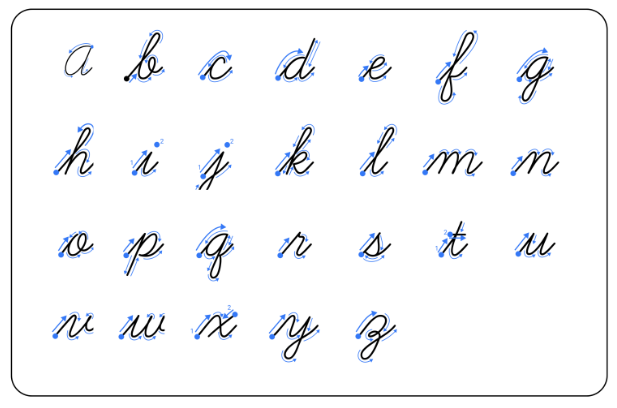}
    \caption{Stroke pattern for British Small cursive that was followed during data collection.}
    \label{fig:placeholder}
\end{figure}

\subsection{Ethics and Reproducibility Statements}

\textbf{Ethics Statement:} All data was collected from consenting adult participants who were informed of the study's purpose. The dataset was fully anonymized, with no personally identifiable information collected or stored. The goal of this research is to advance scientific understanding of AI capabilities and limitations. We do not foresee any direct negative societal impacts, and the dataset does not contain any sensitive content.

\textbf{Reproducibility Statement:} The Unwritten Benchmark dataset has been made publicly available on our Project Page. All models were accessed via their APIs or HuggingFace, and the exact model versions are noted in the main text. The prompts used for evaluation are provided in their entirety in the paper to ensure that our results are fully reproducible.

\end{document}